\documentclass[runningheads]{llncs}
\usepackage[pdftex]{graphicx}
\usepackage{amsmath}
\usepackage{amsfonts}
\usepackage{amssymb}
\usepackage{verbatim}
\usepackage{booktabs}
\usepackage[hidelinks,hypertexnames=true]{hyperref}
\usepackage{subfigure}
\usepackage[inline]{enumitem}

\usepackage{url}
\usepackage{color}

\newcommand{\cf}{{\it cf.}}
\newcommand{\eg}{{\it e.g.}}

\newcommand{\reals}{\mathbf{R}}

\newcommand{\sat}{\texttt{sat}}
\newcommand{\unsat}{\texttt{unsat}}

\newcommand{\one}{\text{`}\texttt{1}\text{'}}
\newcommand{\two}{\text{`}\texttt{2}\text{'}}
\newcommand{\three}{\text{`}\texttt{3}\text{'}}
\newcommand{\four}{\text{`}\texttt{4}\text{'}}
\newcommand{\five}{\text{`}\texttt{5}\text{'}}
\newcommand{\six}{\text{`}\texttt{6}\text{'}}
\newcommand{\seven}{\text{`}\texttt{7}\text{'}}
\newcommand{\eight}{\text{`}\texttt{8}\text{'}}
\newcommand{\nine}{\text{`}\texttt{9}\text{'}}
\newcommand{\zero}{\text{`}\texttt{0}\text{'}}

\newcommand{\PyTorch}{\textsc{PyTorch}}
\newcommand{\Lantern}{\textsc{Lantern}}
\newcommand{\Reluplex}{\textsc{Reluplex}}

\newcommand{\dbias}{\ensuremath{\Delta\theta_{\mathrm{bias}}}}
\newcommand{\dweight}{\ensuremath{\Delta\theta_\mathrm{weight}}}
\newcommand{\dtheta}{\ensuremath{\Delta\theta}}

\newlist{mylist}{enumerate*}{1}
\setlist[mylist]{label=(\roman*)}

\title{Incorrect by Construction: Fine Tuning Neural Networks for Guaranteed Performance on Finite Sets of Examples}
\titlerunning{Incorrect by Construction: Fine Tuning Neural Networks}

\author{Ivan Papusha\inst{1} \and 
        Rosa Wu\inst{1,2} \and 
        Joshua Brul{\'e}\inst{1} \and
        Yanni Kouskoulas\inst{1} \and
        Daniel Genin\inst{1} \and 
        Aurora Schmidt\inst{1}}
\authorrunning{Papusha, et al.}
\institute{Johns Hopkins University Applied Physics Laboratory\thanks{This work
           was supported by JHU/APL Internal Research and Development funds.} \and
           Defense Nuclear Facilities Safety Board\thanks{The views expressed
           herein are solely those of the authors, and no official support or
           endorsement by the Defense Nuclear Facilities Safety Board or the
           U.S. Government is intended or should be inferred.}}

\begin{document}
\maketitle


\begin{abstract}
There is great interest in using formal methods to guarantee the reliability of
deep neural networks. However, these techniques may also be used to implant
carefully selected input-output pairs. 
We present initial results on a novel technique for using SMT solvers to fine
tune the weights of a ReLU neural network to guarantee outcomes on a finite set
of particular examples. This procedure can be used to ensure performance on key
examples, but it could also be used to insert difficult-to-find incorrect
examples that trigger unexpected performance. We demonstrate this approach by 
fine tuning an MNIST network to incorrectly classify a particular image and
discuss the potential for the approach to compromise reliability of
freely-shared machine learning models. 
\end{abstract}


\section{Introduction}
\label{sec:intro}

Advances in the construction and training of deep neural networks have
transformed many problems in classification, machine learning, and autonomous
systems. But the large number of internal degrees of freedom that
make these networks so powerful can also prove to be a source of
vulnerability---verifying that such complex systems always perform in an
expected way is a daunting task. As a result, there is much interest in using
automatic formal verification techniques that employ satisfiability modulo
theories (SMT) to generate guarantees about the behavior of such networks. 

SMT is a recently attractive technology because of practical solver advances and
mature implementations.  Leveraging a complete decision procedure, solvers can
generate a network input that satisfies a given constraint (\sat), or guarantee
that no such input exists (\unsat). By treating perturbations to the network as
variable, we find that we may also use SMT to search for small
modifications to the network itself that guarantee performance it did not
already have. 

In this work, we use Z3~\cite{deMoura:2008} to embed a set of guaranteed
input-output examples by taking advantage of the ample degrees of freedom in
the biases. Our main contribution is to show that small bias perturbations can
internally model intentionally-planted correct or incorrect input-output pairs
with moderately reduced performance on the off-target examples. Our
approach could be used to fine tune networks to guarantee performance on a
critical set of examples, or to poison them with malicious triggers.
Furthermore, the technique is constructive---we either exhibit
specific bias perturbations satisfying prescribed constraints, or generate a
verifiable proof artifact showing that none exist.


\vspace{-1ex}
\paragraph{Prior work}
The rise in effective optimization techniques for producing adversarial
examples has led to an explosion of interest in how to fool neural networks
with inputs that are slight modifications of correctly classified examples.
Much effort has been devoted to finding such adversarial examples in various
neural networks~\cite{GoodfellowSS14,yuan2019adversarial}.
This has inspired researchers to use SMT to verify or construct neural networks
lacking such adversarial examples~\cite{Bohrer:2019,carlini2017provably,Cheng:2017}.

Like \Reluplex{} and related approaches that use SMT to find adversarial
examples or guarantee their
absence~\cite{Katz2017ReluplexAE,Papusha:2018b,Bunel:2017,Ehlers:2017,Huang:2017,Dutta:2019},
we restrict our
attention to neural networks with piecewise affine activation layers. These
include, for example, rectifier linear unit (ReLU) and HardTanh, but
not sigmoid or softmax layers.
However, instead of searching for perturbations
on \emph{inputs}, we search for perturbations on the network \emph{biases},
thereby globally and tractably parameterizing all possible neural networks of a
certain class. 

Although we use MNIST as a running example, our patching approach inherits the
adverse scaling of SMT with neural network size, likely precluding adoption to
vision tasks in the near term. Our philosophy is therefore not just to retrain with
a modified training set as in~\cite{Gu:2017}, but rather to globally and
reliably optimize over the space of all neural networks that satisfy a set of
constraints.  This allows application in broader frameworks for design and
verification of high reliability systems with formal guarantees on end-to-end
behavior~\cite[\S{}2]{Papusha:2018b}.


As part of this paper, we review methods (\S\ref{sec:smt}) for translating a
neural network into SMT constraints, and follow with detail on using the
encoded network to generate adversarial inputs (\S\ref{sec:method}), as well as
adjusting the network parameters to implant guaranteed input-output pairs.
We demonstrate this approach (\S\ref{sec:experiments}) by implanting behavior
in a small example network for digit classification.  We conclude
(\S\ref{sec:conclusion}) with a discussion of the potential for this method to
scale to larger networks, as well as future work. 
\vspace{-2ex}

\section{Neural Network as Constraints}
\label{sec:smt}
%
%

The key insight to our approach is the observation that certain neural networks
are well-suited to analysis via SMT, while still
being expressive enough to perform calculations of interest, see
\eg,~\cite{Papusha:2018b,Bunel:2017}.
We encode the input-output relations of deterministic neural networks as
quantifier free combinations of linear arithmetic constraints.

\subsection{Piecewise Affine Networks}
\label{sec:nn-encoding}
Consider a network $f:\reals^n \to \reals^m$, represented by a function
$y=f(x;\theta)$, with parameters $\theta$. The input $x$ and
output $y$ are $n$- and $m$-dimensional real vectors, respectively.
In a typical architecture, the neural network is designed as a sequential
composition of alternating affine functions and piecewise affine (\eg, ReLU,
HardTanh) activations,
\begin{equation}
	\label{eq:neural-net}
	f = \beta_K \circ \alpha_K \circ \cdots \circ \beta_1 \circ \alpha_1.
\end{equation}

Each affine function $\alpha_k: \reals^{n_k} \to \reals^{m_k}$ is
parameterized by a dense $m_k$-by-$n_k$ real weight matrix
$W^{(k)}$ and an $m_k$-dimensional bias vector $b^{(k)}$,
\begin{align}
	\label{eq:aff-layer}
	[\text{Affine}]   & \quad \alpha_k(x; W^{(k)}, b^{(k)}) := W^{(k)} x + b^{(k)}.
\end{align}
Similarly, the activation functions $\beta_k : \reals^{m_k} \to \reals^{m_k}$
are piecewise affine. We consider componentwise ReLU and HardTanh activations,
although any piecewise affine activation can be likewise treated,
\begin{align}
	\label{eq:relu-layer}
	[\text{ReLU}]     & \quad {\beta_k(x)}_i := \max(x_i, 0), \quad i=1,\ldots,m_k, \\
	\label{eq:hardtanh-layer}
	[\text{HardTanh}] & \quad {\beta_k(x)}_i := \max(\min(x_i, 1), -1), \quad i=1,\ldots,m_k.
\end{align}

The input dimension of the network $f$ is $n = n_1$ in the first
layer, and the output dimension is $m = m_K$ in the last. The
signal dimension can only change in the affine layers ($m_k \neq n_k$ in
general), and remains the same through the activations ($m_k = n_{k+1}$).
For convenience, we split the network parameters into
weight and bias components $\theta = (\theta_\text{weight},
\theta_\text{bias})$,
\begin{align*}
	\theta_\text{weight} &= (W^{(1)},\ldots,W^{(K)}) 
		\in \reals^{m_1\times n_1} \times \cdots \times \reals^{m_K \times n_K}, \\
	\theta_\text{bias}   &= (b^{(1)},\ldots,b^{(K)}) 
		\in \reals^{m_1} \times \cdots \times \reals^{m_K}.
\end{align*}

\subsubsection{SMT encoding}
We encode the neural network by introducing intermediate variables
$x^{(1)},\ldots,x^{(K+1)}, y^{(1)},\ldots,y^{(K)}$ to hold the
results of the compositions in~\eqref{eq:neural-net}. Specifically, for an
input variable $x$ and output variable $y$, the input-output relation of the
neural network $y = f(x;\theta)$ is equivalent to
\begin{equation}
	\label{eq:nn-encoding}
	(x = x^{(1)})
	\wedge
	\left(\bigwedge_{k=1}^{K} x^{(k+1)} = \beta_k(y^{(k)}) \wedge y^{(k)} = \alpha_k(x^{(k)})\right)
	\wedge
	(y = x^{(K+1)}).
\end{equation}
The affine layers are encoded as-is following~\eqref{eq:aff-layer},
\begin{align}
	\label{eq:aff-encoding}
	[\text{Affine-Encoding}]   & \quad
		v = \alpha_k(u) 
		\Longleftrightarrow 
		v = W^{(k)} u + b^{(k)},
\end{align}
with variables $u, v$ and parameters $W^{(k)}, b^{(k)}$.

To encode the activation functions, note that equality constraints
involving `$\min$' and `$\max$' can be written as a logical combination
of affine atoms:
\begin{align}
	\eta = \min(\xi, a) &
	\Longleftrightarrow 
			[(\xi \geq a) \to (\eta = a)] \wedge 
			[(\xi < a) \to (\eta = \xi)],\\
	\eta = \max(\xi, b) &
	\Longleftrightarrow 
			[(\xi < b) \to (\eta = b)] \wedge 
			[(\xi \geq b) \to (\eta = \xi)].
\end{align}
Accordingly, the piecewise affine activation functions are logical conjunctions
over individual components,
\begin{align}
	[\text{ReLU-Encoding}]
		& \quad v = \beta_k(u)
		\Longleftrightarrow
		\bigwedge_{j=1}^{m_k}
			(v_j = \max(u_j, 0)),
		\label{eq:relu-encoding}
		\\
	[\text{HardTanh-Encoding}] 
		& \quad v = \beta_k(u)
		\Longleftrightarrow
		\bigwedge_{j=1}^{m_k}
			(v_j = \max(\min(u_j, 1), -1)).
		\label{eq:hardtanh-encoding}
\end{align}
Put together, equations~\eqref{eq:aff-encoding}--\eqref{eq:hardtanh-encoding}
can be substituted successively into equation~\eqref{eq:nn-encoding}, resulting
in an encoding of the neural network~\eqref{eq:neural-net} into a single
formula consisting of conjunctions, disjunctions, and negations of affine
atoms. Thus, any neural network constraint of the form $y=f(x;\theta)$, with
variables $x$ and $y$, and parameters $\theta$, corresponds to a conjunction of
constraints of the form~\eqref{eq:nn-encoding}.

\subsection{Using Pretrained PyTorch modules}
\label{sec:pytorch}

To automate our experiments, we developed a Python package, \Lantern{}
(``safer than a torch''), which converts common neural network modules
from the popular
\PyTorch{} library~\cite{Paszke:2019} to variables and
constraints that can be further manipulated with an SMT solver such as
Z3~\cite{deMoura:2008}. We assume that the (trained) network is 
represented as a \verb|Sequential| module, a \PyTorch{} container that
holds other modules and applies them in sequence. 
We further assume that the modules within a given
\verb|Sequential| instance are either \verb|Linear|, \verb|ReLU|, or \verb|Hardtanh|. 

For each module in a \verb|Sequential|, \Lantern{} generates Z3 variables that
correspond to the inputs and outputs of that module, and encodes the behavior of
that module as affine constraints (see~\S\ref{sec:nn-encoding}).
In addition, it creates constraints that equate the output variables of each
module with the input variables of the next module in the sequence. This
process returns the input and output variables of the entire \verb|Sequential|,
as well as all the constraints that represent the internal modules.

The default settings of \PyTorch{} result in models parameterized by 32-bit
floats, which can give computationally difficult SMT formulas.  When the floats
are losslessly cast to \verb|Real|-sorted variables, formulas involving
the neural network can be handled using Z3's linear real arithmetic solver.
However, in practice we found that arbitrary precision calculations often
dominated decision run times, meaning that computations involving even
moderately sized networks benefited from a translation to IEEE floating-point
arithmetic. Therefore, our software also supports quantizing networks into
floating-point representations.

The \verb|round_model()| function truncates the significand of the
floating-point parameters of a trained network to a desired number of
bits. This function provides an adjustable trade-off between the neural
network's performance and the difficulty of the corresponding SMT problem. 
A rounded model remains a valid \verb|Sequential| object, and can be run
just like the original at inference time, albeit with reduced accuracy. 
By quantizing the model itself, we preserve a one-to-one correspondence between
the SMT problem and the network, even though the rounded model is no longer
equivalent to the original.

\section{Method for Planting Examples}
\label{sec:method}
%


A common application of the SMT encoding (\S\ref{sec:nn-encoding}) is to find
perturbations on an input that would result in a classifier
misclassifying otherwise correct examples. The existence of
techniques to find small perturbations is well
documented~\cite{Huang:2017,Ehlers:2017}.  We will briefly summarize these
findings (\S\ref{sec:adversarial-input}) with an eye toward explaining our
novel neural network modification strategy (\S\ref{sec:adversarial-nnmod}).

\subsection{Adversarial Input Generation}
\label{sec:adversarial-input}

Consider a trained network $f$, which \emph{correctly} classifies an input 
$x^0$ as $y^0$, so that specifically $y^0 = f(x^0; \theta)$ for the given input-output 
pair $(x^0, y^0)$. We would like the network to instead output a specified $y^1$ 
for a perturbed input $x^0 + \Delta x$, where $y^0 \neq y^1$ and the perturbation 
magnitude $\|\Delta x\|$ is small.


In this setting, finding a minimal adversarial input amounts to solving the
(nonconvex) optimization problem
\begin{equation}
    \label{eq:adv-input-opt}
	\begin{array}{ll}
	  \mbox{minimize} & \|\Delta x\|\\
	  \mbox{subject to} & f(x^0 + \Delta x; \theta) = y^1
	\end{array}
\end{equation}
over the variable $\Delta x\in\reals^n$. This will give a smallest perturbation 
$\Delta x$ on the input that is enough to get the network to misclassify $x^0$ as 
$y^1$. We target the $\ell_\infty$ norm $\|\cdot\|$, 
because it can be represented with piecewise affine (`$\max$') functions, 
although many other norms are possible. 
The parameters of the network $\theta$ remain constant throughout the 
adversarial input generation process. 

\subsubsection{Optimal perturbation}
The objective in~\eqref{eq:adv-input-opt} can be minimized with bisection by 
posing a sequence of queries to the SMT solver. Specifically, define the formula
\[
	F(\alpha) = \exists \Delta x \in\reals^m . \  (y^1 = f(x^0 + \Delta x; \theta)) \wedge (\|\Delta x\| \leq \alpha).
\]

If, for a given value of $\alpha \in [\alpha_-, \alpha_+]$ 
(where $F(\alpha_+)$ is \sat{} and $F(\alpha_-)$ is \unsat), 
the formula $F(\alpha)$ is \sat, then we know that at the optimum $\|\Delta x^\star\| \leq \alpha$; 
we should therefore decrease the upper bound to $\alpha_+ := \alpha$, and 
determine the satisfiability of $F((\alpha_+ + \alpha_-)/2)$, say. 
Otherwise if $F(\alpha)$ is \unsat, then a valid input perturbation 
must have norm no less than $\alpha$; therefore, to make the network misclassify 
$x^0$ as $y^1$ we should increase the lower bound to $\alpha_- := \alpha$, 
and try again. This way, a minimal value of $\|\Delta x\|$ can be determined
within an error $\epsilon$ in $O(\log_2(1/\epsilon))$ bisection steps.

\subsubsection{Class membership}
Encoding the constraint $y^1 = f(x^0 + \Delta x; \theta)$ in~\eqref{eq:adv-input-opt} 
deserves special attention in the case of classifiers, because class membership 
must be encoded with set membership (\eg, lying on the correct side of a decision surface).
For example, for $m=10$ (MNIST digit classification problem), 
we identify the output indices with the classes 
\one{}, \two{}, \ldots, \nine{}, \zero{}.
For example, the network output
\[
	f(x^0; \theta) = \begin{bmatrix}
	  0.01\\
	  0.95\\
	  \ldots\\
	  0.02
	\end{bmatrix}
\]
is interpreted as \two{}, because the second component has maximal value
(softmax layers are disallowed in linear SMT theories).
A class equality constraint like $y = \seven{}$ is in reality a requirement
on the seventh component of $y$ to be maximal, 
\begin{equation}
	\label{eq:class-equality}
	(y_7 > y_1) \wedge
	\cdots \wedge
	(y_7 > y_6) \wedge
	(y_7 > y_8) \wedge
	\cdots \wedge
	(y_7 > y_{10}).
\end{equation}
A class membership constraint is thus a conjunction of affine constraints.

\subsubsection{Forcing correctness}
The same adversarial input generation technique can be used if the correctness
senses of $y^0$ and $y^1$ are switched: when the network \emph{incorrectly}
classifies $x^0$ as $y^0$, then solving the optimization
problem~\eqref{eq:adv-input-opt} is akin to finding a minimum-size
perturbation on the input that will \emph{force} the output to the desired
correct value $y^1$. In this case, the network outputs a correct value with a
small input perturbation, even if it originally failed to do so.

\subsection{Adversarial Network Modification}
\label{sec:adversarial-nnmod}

The idea of forcing output values introduced in the previous section can 
similarly be used to \emph{patch} the network parameters 
to achieve desired
performance on specified input-output pairs. The key difference
lies in patching the biases only, meanwhile keeping the weights fixed.


\subsubsection{Bias patching}
Consider a supervised task with a training database of input-output pairs 
$D = \{(x,y)\} \subset \reals^n \times \reals^m$. We would like 
to keep the neural network output values the same on a finite set 
$D^\text{keep} \subset \reals^n \times \reals^m$, 
and force a change on a finite set 
$D^\text{change} \subset \reals^n \times \reals^m$ 
of values.
It is not necessary that $D^\text{keep}$ or $D^\text{change}$ 
be subsets of $D$, but we require that 
$D^\text{keep} \cap D^\text{change} = \emptyset$.
The procedure for patching the network biases consists of two conceptual steps:
\begin{enumerate}
  \item{} [Train] Classically train (\eg, using stochastic gradient descent) 
  a ReLU network $f(x; \theta)$ on the database $D$, obtaining the parameter 
  vector $\theta = (\theta_\text{weight}, \theta_\text{bias})$ as a starting point.
  \item{} [Patch] Keeping the weight component $\theta_\text{weight}$ fixed, 
  modify the network from Step 1 by solving the optimization problem
  \begin{alignat}{3}
    &\mbox{minimize }   && \|\dtheta\| \quad && \nonumber \\
    &\mbox{subject to } && y = f(x; \theta + \dtheta), \quad && \text{for all } (x,y) \in D^\text{keep}, \label{eq:mod-a} \\
    & && y' = f(x'; \theta + \dtheta), \quad && \text{for all } (x', y') \in D^\text{change}, \label{eq:mod-b} \\
    & && \dweight = 0. \label{eq:mod-c}
  \end{alignat}
  over the variables $\dtheta = (\dweight, \dbias)$.
\end{enumerate}

Classical neural network training will not
(typically) result in a parameter vector $\theta$ that correctly assigns 
all points in $D$. However, the SMT patching procedure will force the values in 
$D^\text{keep}$ and $D^\text{change}$, or otherwise return a proof
that a network modification of the prescribed type is impossible.

\subsubsection{Linear arithmetic}
The biases can can be patched because they enter affinely into the neural network 
constraints~\eqref{eq:nn-encoding} (whereas the weights enter multiplicatively).
As a result, bias perturbation variables can be added at each 
$\alpha_k$ network layer while still using a decision procedure based on 
linear arithmetic, \cf~\eqref{eq:aff-encoding},
\begin{equation}
	\label{eq:aff-encoding-perturb}
		v = \alpha_k(u) 
		\Longleftrightarrow 
		v = W^{(k)} u + b^{(k)} + \dbias^{(k)}.
\end{equation}
Staying within a linear decision theory helps performance, 
although we expect weight modification with nonlinear theories 
(and multiplicative terms) to be practical in small networks~\cite{Gao:2012}.

A key scaling challenge lies in keeping the fewest number of constraints
in~\eqref{eq:mod-a} and~\eqref{eq:mod-b}, since there are as many instances of
the fully encoded neural network in the optimization problem as there are
examples in $D^\text{keep}\cup D^\text{change}$. To help this potential
difficulty, it is desirable to keep $|D^\text{keep}|$ and $|D^\text{change}|$
small.

\section{Experiments}
\label{sec:experiments}
Following the outlined approach (\S\ref{sec:method}), we encoded small- and
medium-sized neural networks to test the generation of adversarial inputs in
realistic cases. Additionally, we modified the medium-sized network to give
prescribed outputs for prescribed inputs.  We performed experiments using the
MNIST database of handwritten digits~\cite{lecun-mnisthandwrittendigit-2010}. 

Because the computational complexity of the SMT decision procedure is heavily
dependent on the total number of units in the network under consideration, the
28-by-28 pixel grayscale images of handwritten digits were flattened to vectors
of length 784, and dimensionally reduced with Principal Component Analysis
(PCA) by selecting the top-30 or top-100 components, depending on the
experiment.  To improve runtime of the solver, network weights and biases were
rounded using the \verb|round_model()| function (\S\ref{sec:pytorch}).



\subsection{Adversarial Input Generation}
The ``small'' MNIST classifier architecture is shown below.
\begin{equation*}
    \label{eq:mnist_neural_net1}
    [\text{image}]
    \overset{784}{\rightarrow}
    \fbox{\strut PCA}
    \overset{30}{\rightarrow}
    \underbrace{\left[
    [x]
    \overset{30}{\rightarrow}
    \fbox{\strut Linear}
    \overset{10}{\rightarrow}
    \fbox{\strut ReLU}
    \overset{10}{\rightarrow}
    \fbox{\strut Linear}
    \overset{10}{\rightarrow}
    \fbox{\strut ReLU}
    \overset{10}{\rightarrow}
    [y]
    \right]}_{y=f(x;\theta)}
\end{equation*}
For the first experiment, the top-30 principal component network was probed to
see if there exist adversarial inputs to make the network misclassify specific
images.
The image representations with reduced dimensionality are treated as inputs
to~$f$,
a four-layer \PyTorch{} \verb|Sequential| model composed of alternating
\verb|Linear| and \verb|ReLU| modules trained with stochastic gradient descent.
The components of the output vector are used to decide the digit
class according to~\eqref{eq:class-equality}.
This small network achieves 72.0\% accuracy on the validation data.

\begin{figure}[htbp]
    \vspace{-10pt}
    \centering
    \includegraphics[width=.9\textwidth]{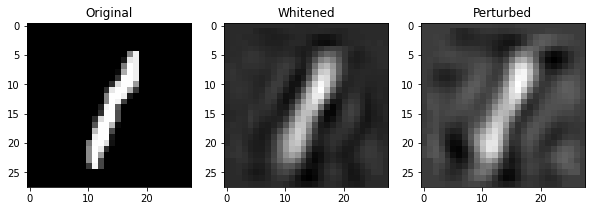}
    \caption{The network correctly classifies the original MNIST image (left) 
             as~\one{} by observing the top-30 PCA components (middle).
             The input-perturbed image is misclassified as~\seven{} (right).}
    \label{fig:one_to_seven}
    \vspace{-15pt}
\end{figure}


We use \Lantern{} to encode the small network as Z3 constraints.
Then we find a reduced-dimensionality representation of an image of a~\one{}
that the network can correctly classify, and force that image to misclassify
as~\seven{}.
Figure~\ref{fig:one_to_seven} shows the original image, the image after PCA compression, 
and finally the misclassified version with an adversarial perturbation 
having magnitude $\|\Delta x\|_\infty = 0.4$.

\subsection{Adversarial Network Modification}
In this experiment, our goal was to modify the network biases such that several
\one{} instances would be misclassified as~\seven{}, while the other classes
continued to be accurately classified. To test scalability and network
quantization, we used a slightly larger, top-100 component PCA compressed data
set with a similar neural network architecture:
\begin{equation*}
    \label{eq:mnist_neural_net2}
    [\text{image}]
    \overset{784}{\rightarrow}
    \fbox{\strut PCA}
    \overset{100}{\rightarrow}
    \underbrace{\left[
    [x]
    \overset{100}{\rightarrow}
    \overset{\begin{gathered}+\dbias \\ \downarrow \end{gathered}}{\fbox{\strut Linear}}
    \overset{33}{\rightarrow}
    \fbox{\strut ReLU}
    \overset{33}{\rightarrow}
    \overset{\begin{gathered}+\dbias \\ \downarrow \end{gathered}}{\fbox{\strut Linear}}
    \overset{10}{\rightarrow}
    \fbox{\strut ReLU}
    \overset{10}{\rightarrow}
    [y]
    \right]}_{y=f(x; \theta + \Delta\theta)}
\end{equation*}
Prior to any bias modifications, this medium-sized network had a 93.0\% overall
classification accuracy. Table~\ref{tab:class_accuracies} breaks down the
accuracy by each digit.

We encoded the linear layers with constraints of the form
\begin{equation}
    \quad y^{(k)} = W^{(k)} x^{(k)} + b^{(k)} + \dbias^{(k)}
    \label{eq:bias_constraint}
\end{equation}
where $W^{(k)}$ and $b^{(k)}$ are the (fixed) network weights and biases for
the layer, $x^{(k)}$ and $y^{(k)}$ are the (variable) inputs and outputs of
each layer, and $\dbias^{(k)}$ are (variable) bias perturbations.


To construct $D^\text{keep}$ and $D^\text{change}$, we chose one set of digits
\zero{} through \nine{}, which the original network classified correctly, and
added them to $D^\text{keep}$. This resulted in $|D^\text{keep}| = 10$
constraints of type~\eqref{eq:mod-a}.
We also found a specific~\one{} image and set its classification target to~\seven{}. 
This resulted in $|D^\text{change}|=1$ constraint of type~\eqref{eq:mod-b}.
Additionally, we added constraints $\|\dbias{}\|_\infty \leq 0.25$
to bound maximum parameter perturbations.


\subsubsection{Guarantees with inherited performance}
The results of our experiment are summarized in
Table~\ref{tab:class_accuracies}, which shows the digit classification
performance of the modified network. The accuracy shown for the modified
network (second column) is an average of two different $D^\text{keep}$ and
$D^\text{change}$ sets. These results indicate a considerable decrease in
accuracy when classifying \one{}s, due to the forced prescription in
$D^\text{change}$, along with moderately smaller accuracy differences in the
other classes.  Note that for the modified network, the eleven examples in
$D^\text{keep}$ and $D^\text{change}$ are guaranteed to be at their prescribed,
forced values.

\begin{table}[htb!]
	\centering
    \caption{Classification accuracy of each digit with the original network 
    weights and biases, modified biases, and modified biases using a quantized model.}
    \label{tab:class_accuracies}
    \begin{tabular}{cccc}
    \toprule
	Digit & \multicolumn{3}{c}{Accuracy (\%)}\\
	\midrule
	{}  & Original & Modified & Quantized\\
	\one{}     & 97.6  & 73.9  & 68.9\\ 
	\two{}     & 92.4  & 89.6  & 91.1\\
	\three{}   & 91.1  & 89.3  & 78.4\\
	\four{}    & 94.0  & 84.2  & 86.1\\
	\five{}    & 86.8  & 77.6  & 91.4\\
	\six{}     & 95.1  & 90.5  & 93.3\\
	\seven{}   & 91.6  & 95.0  & 93.4\\
	\eight{}   & 91.1  & 85.9  & 74.8\\
	\nine{}    & 91.8  & 92.5  & 89.2\\
	\zero{}    & 97.9  & 96.7  & 98.6\\
	\midrule
	Overall & 93.0 & 87.4 & 89.2\\
    \bottomrule
    \end{tabular}
    \vspace{-5pt}
\end{table} 
It takes four hours for Z3 to find a satisfying assignment to \dbias{} in
the top-100 network using linear rational arithmetic.\footnote{Tested on an
Intel(R) Core(TM) i9-8950HK CPU @ 2.90GHz on a 64-bit Windows operating system
with 32.0 GB installed RAM.}
After quantizing the network parameters to 10 bits with \verb|round_model()|,
run times went down to $30$ minutes per floating point arithmetic decision call
($8\times$ speedup).
Performance of the quantized model is shown in the last column of 
Table~\ref{tab:class_accuracies}.
The per-class accuracy for the quantized model is an average of three
different $D^\text{keep}$ and $D^\text{change}$ sets.

\begin{figure}[htb!]
    \vspace{-10pt}
    \centering
    \subfigure[original biases]{
        \label{fig:histograms-a}
        \includegraphics[width=.34\textwidth]{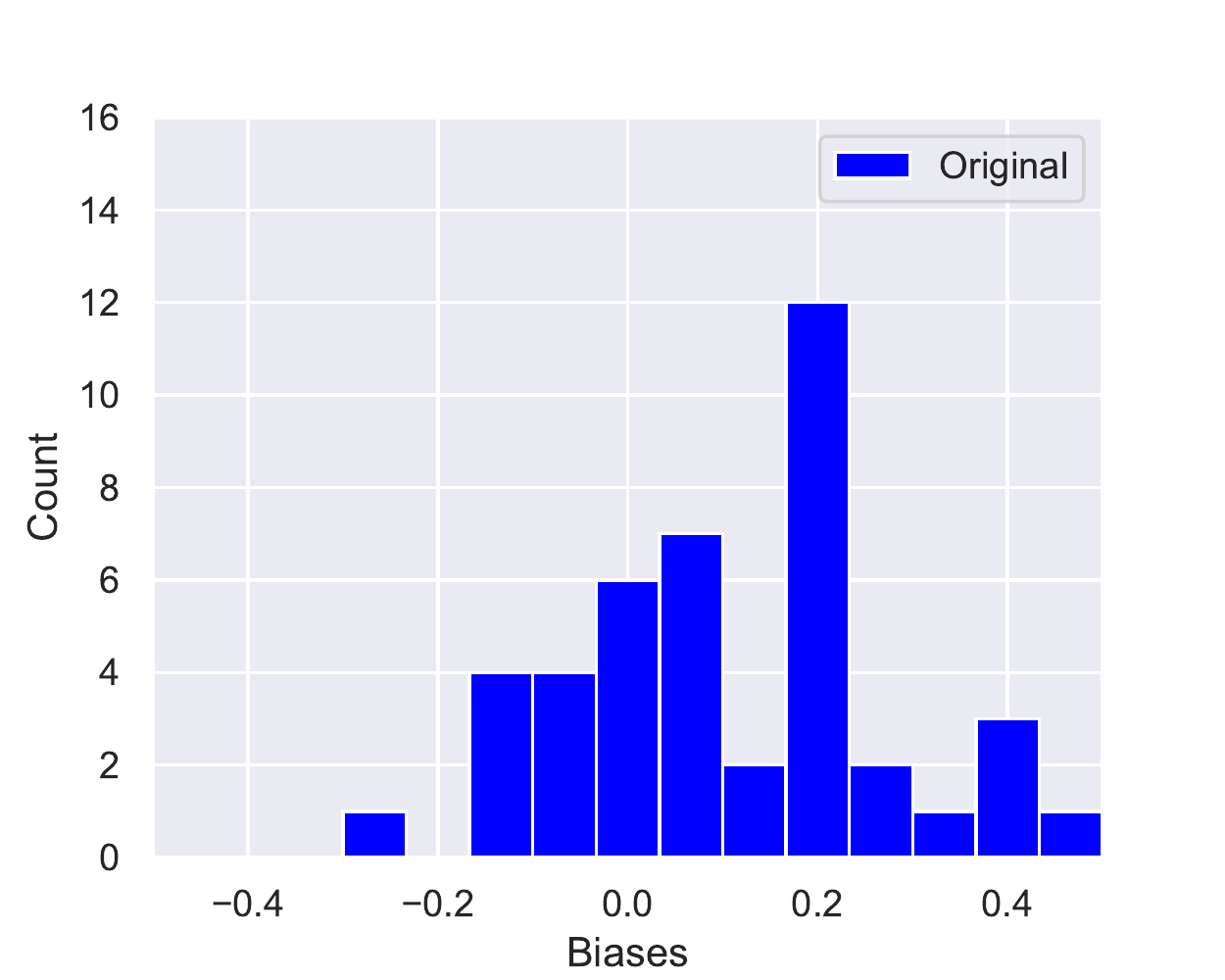}
    }\hspace{-17pt}
    \subfigure[\dbias{} values]{
        \label{fig:histograms-b}
        \includegraphics[width=.34\textwidth]{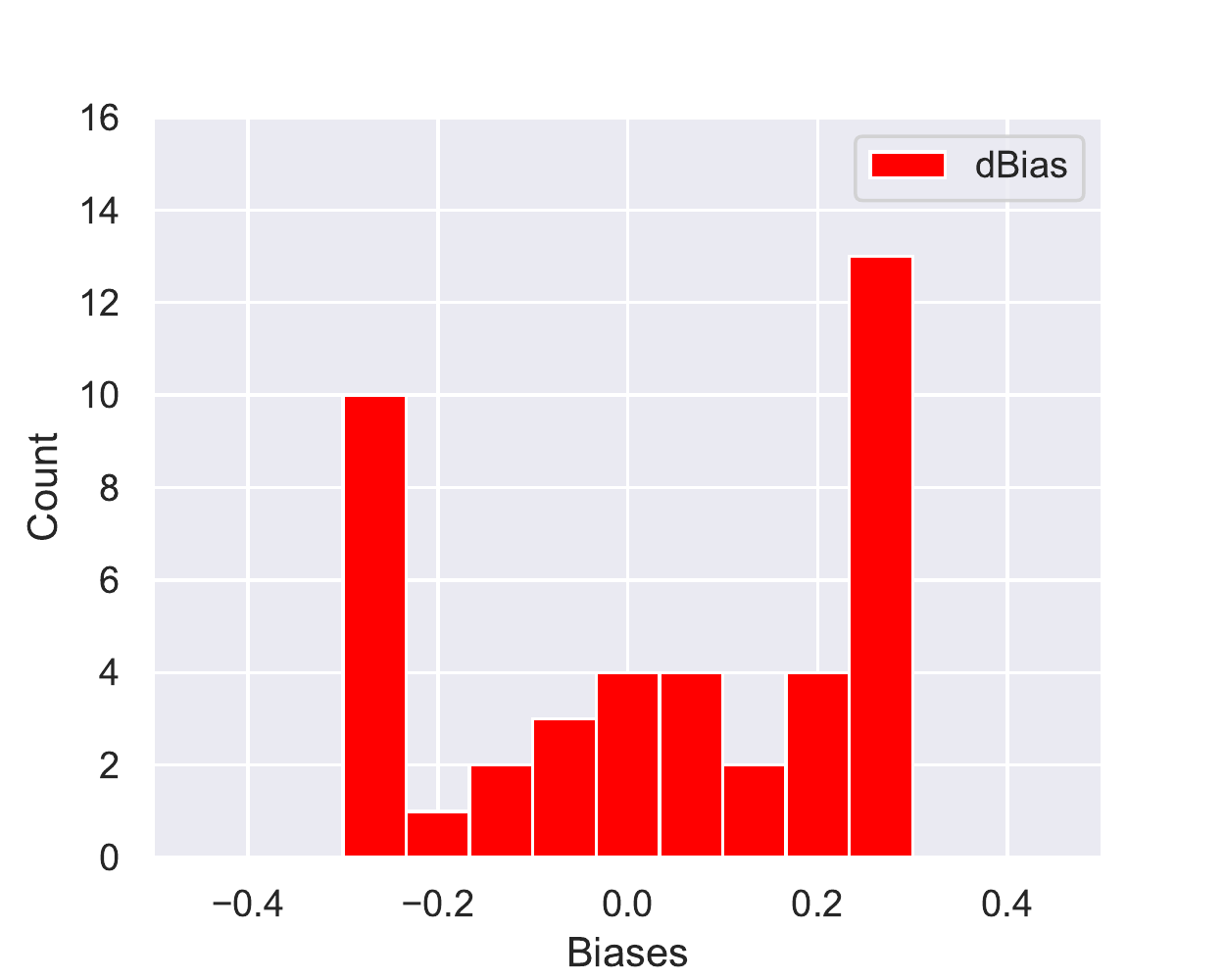}
    }\hspace{-17pt}
    \subfigure[fine-tuned biases]{
        \label{fig:histograms-c}
        \includegraphics[width=.34\textwidth]{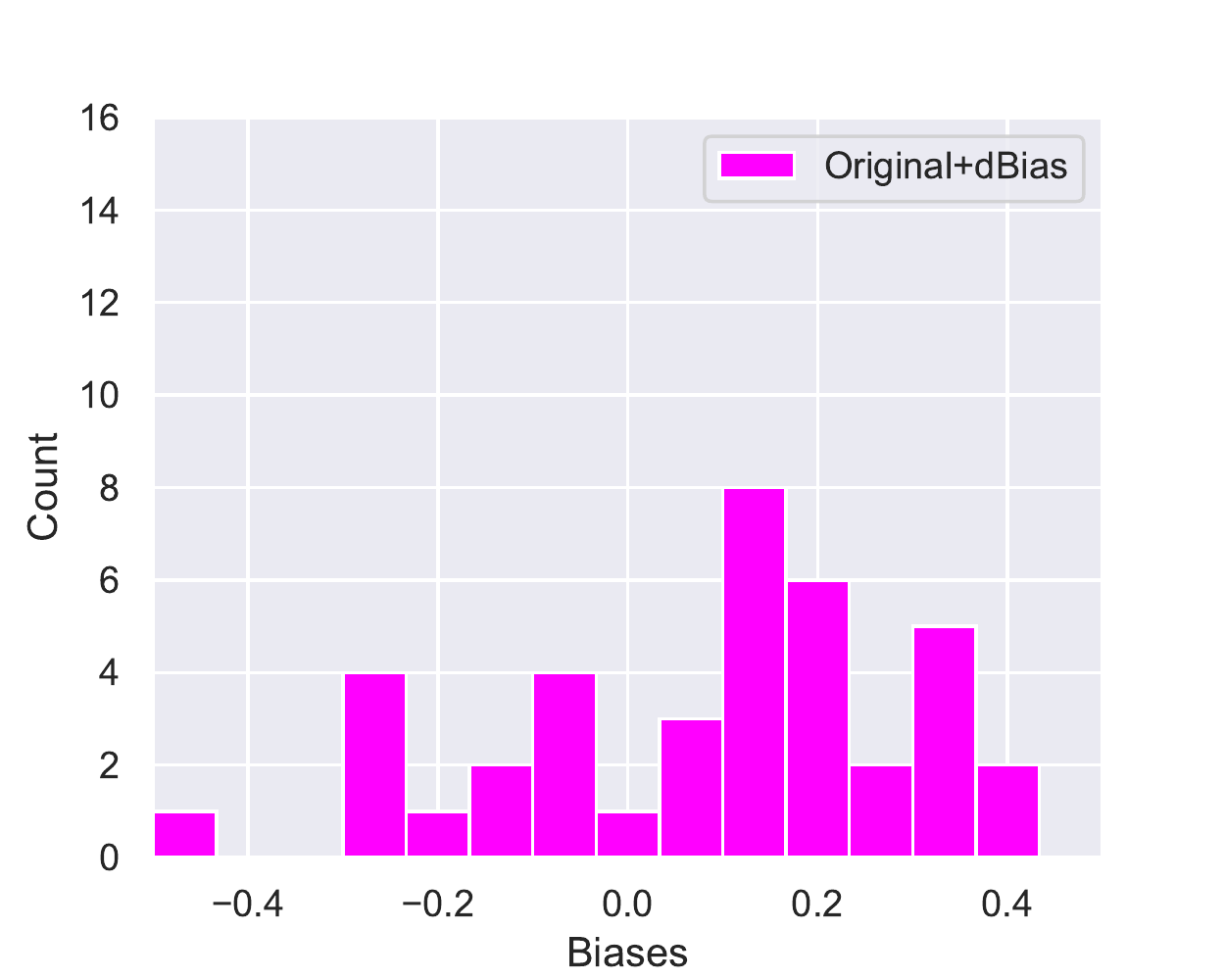}
    }
    \caption{Histograms of \subref{fig:histograms-a} original network biases, 
    \subref{fig:histograms-b} solutions for the bias perturbations for adversarial 
    network modification, and \subref{fig:histograms-c} final modified biases.}
    \label{fig:histograms}
    \vspace{-10pt}
\end{figure}

A visualization of how much the network biases changed is shown in
Figure~\ref{fig:histograms}.
Because the size distributions of the original and perturbed biases are so
similar, Figure~\ref{fig:histograms} suggests that detection of this type of
network tampering may be difficult. Thus, an across-the-board small change in
network biases guaranteed the eleven specific examples to be classified in a
user-prescribed way.


\section{Conclusion and Future Work}
\label{sec:conclusion}

In this work, we have shown how to use SMT to implant behaviors in neural
networks that use piecewise affine activations. In doing so, we also detailed a
method for automatically encoding \PyTorch{} networks into Z3 constraints.
We computed bias perturbations for a relatively small neural network
that performs classification on the MNIST data set. 
We plan to extend this approach to larger neural networks, such as
deep convolutional networks for image recognition and deep reinforcement
learning networks.
In many deep networks, for a particular input, only a small fractions of
neurons end up contributing to the output~\cite{glorot2011deep,Tjeng:2019}.
Thus we can
\begin{mylist}
  \item select a subset of neurons to modify in a large network, and 
  \item improve the efficiency of the decision procedure,
\end{mylist}
by abstracting the majority of the network and locally optimizing the biases of
selected neurons. 

We plan to test a solver algorithm better
optimized to the specific constraint solution problem, thereby increasing the
scale of networks that can be modified and whose performance can be guaranteed. 
Because there are a wealth of networks of modest size, especially those that
perform simple autonomous control, we see value to this approach
despite SMT's unfavorable computational scaling.
Further research on this technique will aid both in understanding the
pitfalls of downloading and using freely shared pretrained neural
networks, as well as the potential for verifying the provable reliability of
neural networks in the loop.
\subsubsection{Acknowledgments}
The authors would like to thank Dr.\ Kiran Karra and Mr.\ Chace Ashcraft for
discussions on this topic and constructive comments on the approach.


\nocite{Papusha:2018b} 
\nocite{Paszke:2019}   

\nocite{Knuth:2015}    
\nocite{Kroening:2016} 
\nocite{Gu:2017}       

\bibliographystyle{splncs04}
\bibliography{refs}

\end{document}